\documentclass[letterpaper, 10 pt, conference]{ieeeconf}  
\usepackage{float} 
\usepackage{times}

\usepackage{multicol}

\usepackage{pdfsync}
\usepackage[dvipsnames,table]{xcolor}
\usepackage{mathtools}
\usepackage{amsmath} 
\usepackage{amssymb}  
\usepackage{multirow}
\usepackage{amsmath}
\usepackage{nccmath} 
\usepackage{algorithm}
\usepackage{pbox}

\makeatletter
\let\NAT@parse\undefined
\makeatother
\usepackage[bookmarks=true]{hyperref}

\newcommand{\separator}{ \noindent \rule{\columnwidth}{1pt} }

\newcommand{\doubleBlind}[1]{} 

\newfloat{spec}{thb}{lop} 
\floatname{spec}{Specification}
\makeatletter
\newcommand{\leqnomode}{\tagsleft@true}
\newcommand{\reqnomode}{\tagsleft@false}
\makeatother

\usepackage{graphicx}
\IEEEoverridecommandlockouts
\begin{document}

\bstctlcite{IEEEexample:BSTcontrol} 

\title{Perception-Informed Autonomous Environment Augmentation With Modular Robots}
\author{
        Tarik~Tosun\textsuperscript{*}, 
        Jonathan~Daudelin\textsuperscript{*}, 
        Gangyuan~Jing\textsuperscript{*}, 
        Hadas~Kress-Gazit, 
        Mark~Campbell, 
        and~Mark~Yim, 
\thanks{J.Daudelin, G. Jing, M. Campbell, and H. Kress-Gazit are with the Sibley School of Mechanical and Aerospace Engineering, Cornell University, Ithaca,
NY, 14850.}
\thanks{T. Tosun and M. Yim are with the Mechanical Engineering and Applied Mechanics Department, University of Pennsylvania, Philadelphia, PA, 19104.}
\thanks{\textsuperscript{*}T. Tosun, J. Daudelin, and G. Jing  contributed equally to this work.}}

\maketitle

\begin{abstract}

We present a system enabling a modular robot to autonomously build structures in order to accomplish high-level tasks. Building structures allows the robot to surmount large obstacles, expanding the set of tasks it can perform.  This addresses a common weakness of modular robot systems, which often struggle to traverse large obstacles.

This paper presents the hardware, perception, and planning tools that comprise our system. An environment characterization algorithm identifies features in the environment that can be augmented to create a path between two disconnected regions of the environment. Specially-designed building blocks enable the robot to create structures that can augment the environment to make obstacles traversable. A high-level planner reasons about the task, robot locomotion capabilities, and environment to decide if and where to augment the environment in order to perform the desired task.  We validate our system in hardware experiments.

\end{abstract}

\IEEEpeerreviewmaketitle


\section{Introduction} \label{sec:introduction}
Employing structures to accomplish tasks is a ubiquitous part of the human experience: to reach an object on a high shelf, we place a ladder near the shelf and climb it, and at a larger scale, we construct bridges across wide rivers to make them passable.  The fields of collective construction robotics and modular robotics offer  examples of systems that can construct and traverse structures out of robotic or passive elements \cite{romanishin20153d,petersen2011termes,terada2004automatic,napp2014distributed}, and assembly planning algorithms that allow arbitrary structures to be built under a variety of conditions \cite{Seo2013,Werfel2007}.  This existing body of work provides excellent contributions regarding the generality and completeness of these methods: some algorithms are provably capable of generating assembly plans for arbitrary volumetric structures in 3D, and hardware systems have demonstrated the capability to construct a wide variety of structures. 

Less work is available regarding ways that robots could deploy structures as a means of completing an extrinsic task, the way a person might use a ladder to reach a high object. In this paper, we present hardware, perception, and high-level planning tools that allow structure-building to be deployed by a modular robot to address high-level tasks.

Our work uses the SMORES-EP modular robot \cite{tosun2016design}, and introduces novel passive block and wedge modules that SMORES-EP can use to form ramps and bridges in its environment.  Building structures allows SMORES-EP to surmount large obstacles that would otherwise be very difficult or impossible to traverse, and therefore expands the set of tasks the robot can perform.  This addresses a common weakness of modular robot systems, which often struggle with obstacles much larger than a module.

We expand on an existing framework for selecting appropriate robot morphologies and behaviors to address high-level tasks \cite{JingAURO2017}. In this work, the high-level planner not only decides when to reconfigure the robot, but also when to augment the environment by assembling a passive structure.
To inform these decisions, we introduce a novel environment characterization algorithm that identifies candidate features where structures can be deployed to advantage.
Together, these tools comprise a novel framework to automatically identify when, where, and how the robot can augment its environment with a passive structure to gain advantage in completing a high-level task.  

We integrate our tools into an existing system for perception-driven autonomy with modular robots \cite{daudelin2017integrated}, and validate them in two hardware experiments. Based on a high-level specification, a modular robot reactively identifies inaccessible regions and autonomously deploys ramps and bridges to complete locomotion and manipulation tasks in realistic office environments.


\section{Related Work}\label{sec:related-work}
%
Our work complements the well-established field of collective robotic construction, which focuses on autonomous robot systems for building activity. While we use a modular robot to create and place structures in the environment, our primary concern is not assembly planning or construction of the structure itself, but rather its appropriate placement in the environment to facilitate completion of an extrinsic high-level task.

Petersen et al. present Termes \cite{petersen2011termes}, a termite-inspired collective construction robot system that creates structures using blocks co-designed with a legged robot. Similarly, our augmentation modules are designed to be easily carried and traversed by SMORES-EP. Where the TERMES project focused on collective construction of a goal structure, we are less concerned with efficient building of the structure itself and more concerned with the application and placement of the structure in the larger environment as a means of facilitating a task unrelated to the structure itself.

Werfel et al. present algorithms for environmentally-adaptive construction that can build around obstacles in the environment \cite{Werfel2007}. A team of robots senses obstacles and builds around them, modifying the goal structure if needed to leave room for immovable obstacles.  An algorithm to build enclosures around preexisting environment features is also presented. As with Termes, the goal is the structure itself; while the robots do respond to the environment, the structure is not built in response to an extrinsic high-level task.

Napp et al. present hardware and algorithms for building amorphous ramps in unstructured environments by depositing foam with a tracked mobile robot \cite{napp2014robotic,napp2014distributed}. Amorphous ramps are built in response to the environment to allow a small mobile robot to surmount large, irregularly shaped obstacles.  Our work is similar in spirit, but places an emphasis on autonomy and high-level locomotion and manipulation tasks rather than construction.

Modular self-reconfigurable robot (MSRR) systems are comprised of simple repeated robot elements (called \emph{modules}) that connect together to form larger robotic structures. These robots can \emph{self-reconfigure}, rearranging their constituent modules to form different morphologies, and changing their abilities to match the needs of the task and environment \cite{Yim2007a}.
Our work leverages recent systems that integrate the low-level capabilities of an MSRR system into a design library, accomplish high-level user-specified tasks by synthesizing library elements into a reactive state machine \cite{JingAURO2017}, and operate autonomously in unknown environments using perception tools for environment exploration and characterization \cite{daudelin2017integrated}.

Our work extends the SMORES-EP hardware system by introducing passive pieces that are manipulated and traversed by the modules. Terada and Murata \cite{terada2004automatic}, present a lattice-style modular system with two parts, structure modules and an assembler robot. Like many lattice-style modular systems, the assembler robot can only move on the structure modules, and not in an unstructured environment. Other lattice-style modular robot systems create structures out of the robots themselves. M-blocks \cite{romanishin20153d} form 3D structures out of robot cubes which rotate over the structure.  Paulos et al. present rectangular boat robots that self-assemble into floating structures, like a bridge \cite{Paulos2015}.

Magnenat et al \cite{magnenat2012autonomous} present a system in which a mobile robot manipulates specially designed cubes to build functional structures. The robot explores an unknown environment, performing 2D SLAM and visually recognizing blocks and gaps in the ground. Blocks are pushed into gaps to create bridges to previously inaccessible areas. In a ``real but contrived experimental design'' \cite{magnenat2012autonomous}, a robot is tasked with building a three-block tower, and autonomously uses two blocks to build a bridge to a region with three blocks, retrieving them to complete its task. Where the Magnenat system is limited to manipulating blocks in a specifically designed environment, our work presents hardware, perception, and high-level planning tools that are more general, providing the ability to complete high-level tasks involving locomotion and manipulation in realistic human environments.

\section{Approach}

\subsection{Environment Characterization}
\label{sec:characterization}
To successfully navigate its environment, a mobile robot must identify traversable areas.
One simple method for wheeled robots is to select flat areas large enough for the robot to fit. However, MSRR systems can reconfigure to traverse a larger variety of terrains. The augmentation abilities we introduce extend MSRR navigation even further; the robot can build structures to traverse otherwise-impossible terrains. For autonomous operation, we need an algorithm to locate and label features in the environment that can be augmented. We present a probabilistic, template-based environment characterization algorithm that identifies augmentable features from a 2.5D elevation map of the robot's environment.

The characterization algorithm searches for a desired feature template $\mathcal{F}_n$ which identifies candidate locations in the environment where useful structures could be built.
A template consists of a grid of likelihood functions $l_i(h)$ for $1 \le i \le M$ where $M$ is the number of grid cells in the template, and $h$ is a height value. The size of grid cells in the template is variable and need not correspond to the resolution of the map. In addition, features of different size can be searched for by changing the cell size of the template to change the scale. In our system implementation, template parameters and likelihood functions are designed by hand to correspond to each structure in the system's structure library. However, future implementations could automatically generate these templates offline with an additional algorithm.

\begin{figure}
\centering
\includegraphics[width=0.45\columnwidth]{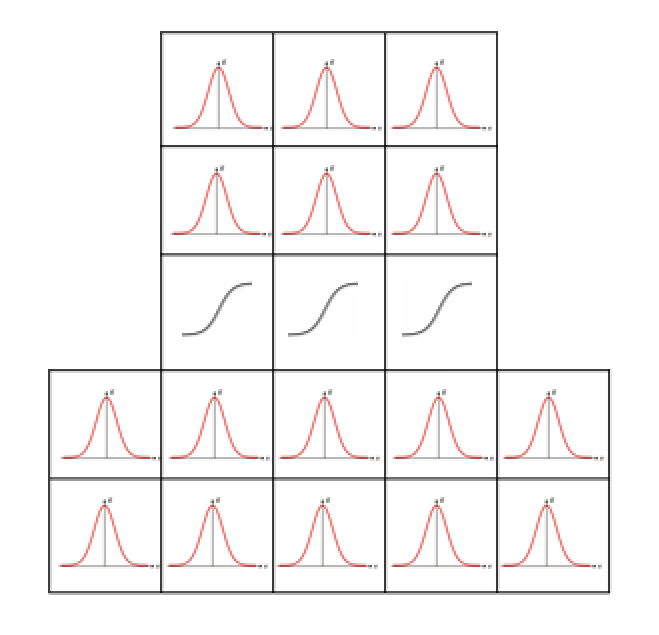}
\includegraphics[width=0.45\columnwidth]{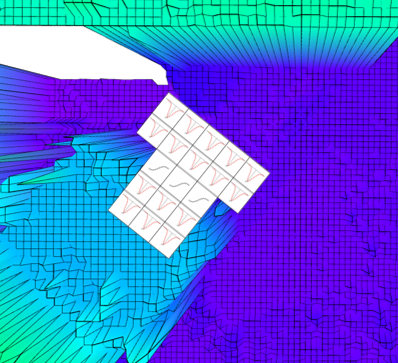}
\caption{\textit{Left:} Example template used to characterize a ``ledge'' feature. \textit{Right:} Example template overlayed on elevation map (top view) to evaluate candidate feature pose.}
\label{fig:template}
\end{figure}

Figure~\ref{fig:template} shows an example of a template used to characterize a ``ledge'' feature, consisting of Gaussian and logistic likelihood functions. Any closed-form likelihood function may be used for each grid cell, enabling templates to accommodate noisy data and variability in possible geometric shapes of the same feature. To determine if the feature exists at a candidate pose $\mathcal{X}$ in the map, a grid of height values is taken from the map corresponding to the template grid centered and oriented at the candidate pose, as illustrated in Figure~\ref{fig:template}. Then, the probability that each grid cell $c_i$ belongs to the feature is evaluated using the cell's likelihood function from the template.

\begin{equation}
P(c_i \in \mathcal{F}_n) = l_i(h_i)
\end{equation}

The likelihood of the feature existing at that location is calculated by finding the total probability that all grid cells belong to the feature. Making the approximate simplifying assumption that grid cells are independent, this probability is equivalent to taking the product over the feature likelihoods of all grid cells in the template:

\begin{equation}
P(\mathcal{X} \in \mathcal{F}_n) = \prod_{i = 1:M} l_i(h_i)
\end{equation}

The feature is determined to exist if the total probability is higher than a user-defined threshold, or $P(\mathcal{X} \in \mathcal{F}_n) > \alpha^M$, where $\alpha$ represents the minimum average probability of each grid cell forming part of the feature. In our experiments we use $\alpha = 0.95$. This formulation normalizes the threshold with respect to the number of grid cells in the template.

\begin{figure}
\begin{center}
\includegraphics[width=0.35\textwidth]{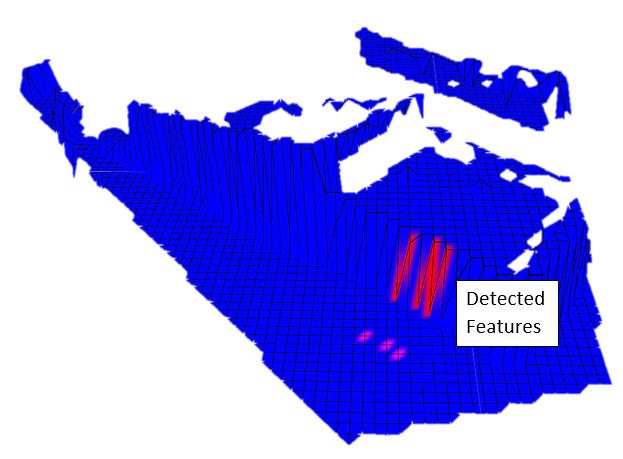}
\caption{Characterization of an environment with a ``ledge'' feature. Red indicates a detected feature, pink indicates the start of the feature, demonstrating orientation.}
\label{fig:characterization} 
\end{center}
\vspace{-3em}
\end{figure}

To characterize an environment, the algorithm takes as inputs an elevation map of the environment and a list of feature templates. Before searching for features, the algorithm preprocesses the elevation map by segmenting it into flat, unobstructed regions that are traversable without augmentation. It then grids the map and exhaustively evaluates each candidate feature pose from the grid, using a grid of orientations for each 2D location. In addition to evaluation with the template, candidate poses are only valid if the ends of the feature connect two traversable regions from the preprocessing step, thereby having potential to extend the robot's reachable space. Once the search is complete, the algorithm returns a list of features found in the map, including their locations, orientations, and the two regions they link in the environment. Figure~\ref{fig:characterization} shows an example of a characterized map. Each long red cell represents a detected ``ledge-height-2'' feature, with a corresponding small pink cell demonstrating the orientation of the feature (and the bottom of the ledge). Note that, in this example, several features are chosen close to each other. Since all connect the same regions, any one is valid and equivalent to be selected for augmentation.

The algorithm scales linearly with the number of grid cells in the 2D environment map, and linearly with the number of features being searched for. Characterization of the environment shown in Figure~\ref{fig:characterization} took approximately 3 seconds to run on a laptop with an Intel Core i7 processor.


%
%
%
%

\subsection{Hardware: Augmentation Modules} 
\label{sec:hardware}
%
Our system is built around the SMORES-EP modular robot. Each module is the size of an 80mm cube, weighs 473g, and has four actuated joints, including two wheels that can be used for differential drive on flat ground \cite{tosun2016design}, \cite{tosun2017paintpots}.  Electro-permanent (EP) magnets allow any face of one module to connect to any face of another, enabling the robot to self-reconfigure. They are also able to attach to objects made of ferromagnetic materials (e.g. steel). The EP magnets require very little energy to connect and disconnect, and no energy to maintain their attachment force of 90N \cite{tosun2016design}. Each module has its own battery, microcontroller, and WiFi module for communication.  In this work, clusters of modules are controlled by a central computer running a Python program that commands movement and magnet via WiFi. Wireless networking is provided by a standard off-the-shelf  router, and commands to a single module can be received at a rate of about 20hz. Battery life is about one hour (depending on magnet, motor, and radio usage).
\begin{figure}   
\begin{center}
\includegraphics[height=1.5in]{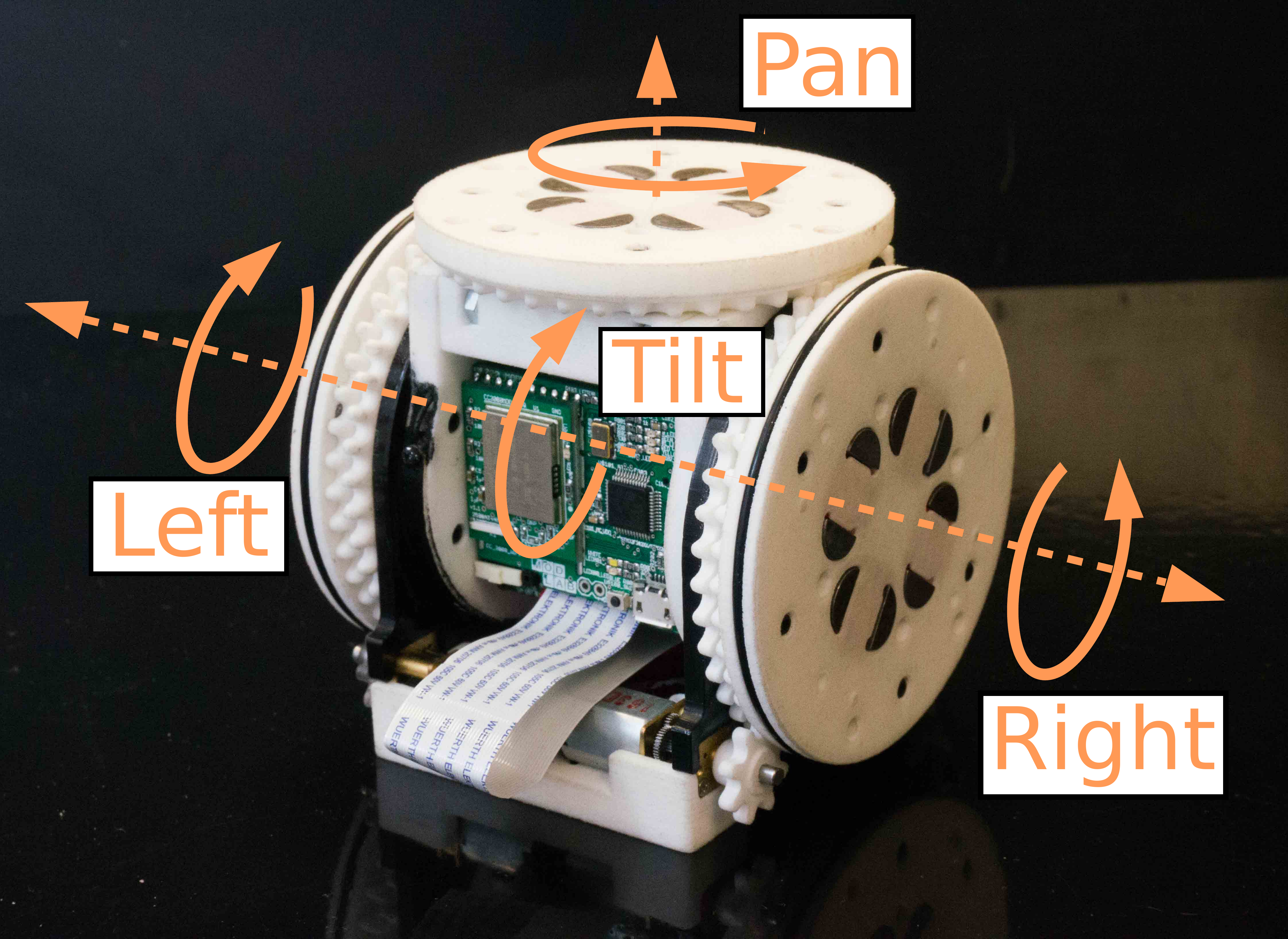}
\end{center}
\caption{SMORES-EP module}
\label{fig:smores-module}
\vspace{-3em}
\end{figure}

Large obstacles, like tall ledges or wide gaps in the ground, are often problematic for modular robot systems. 
One might expect that a modular system could scale, addressing a large-length-scale task by using many modules to form a large robot.  In reality, modular robots don't scale easily: adding more modules makes the robot bigger, but not stronger.  The torque required to lift a long chain of modules grows quadratically with the number of modules, quickly overloading the maximum torque of the   first module in the chain. Consequently, large systems become cumbersome, unable to move their own bodies. Simulated work in reconfiguration and motion planning has demonstrated algorithms that handle hundreds of modules, but in practice, fixed actuator strength has typically limited these robots to configurations with fewer than 40 modules.
%

\begin{figure}
\begin{center}
\includegraphics[width=0.45\columnwidth]{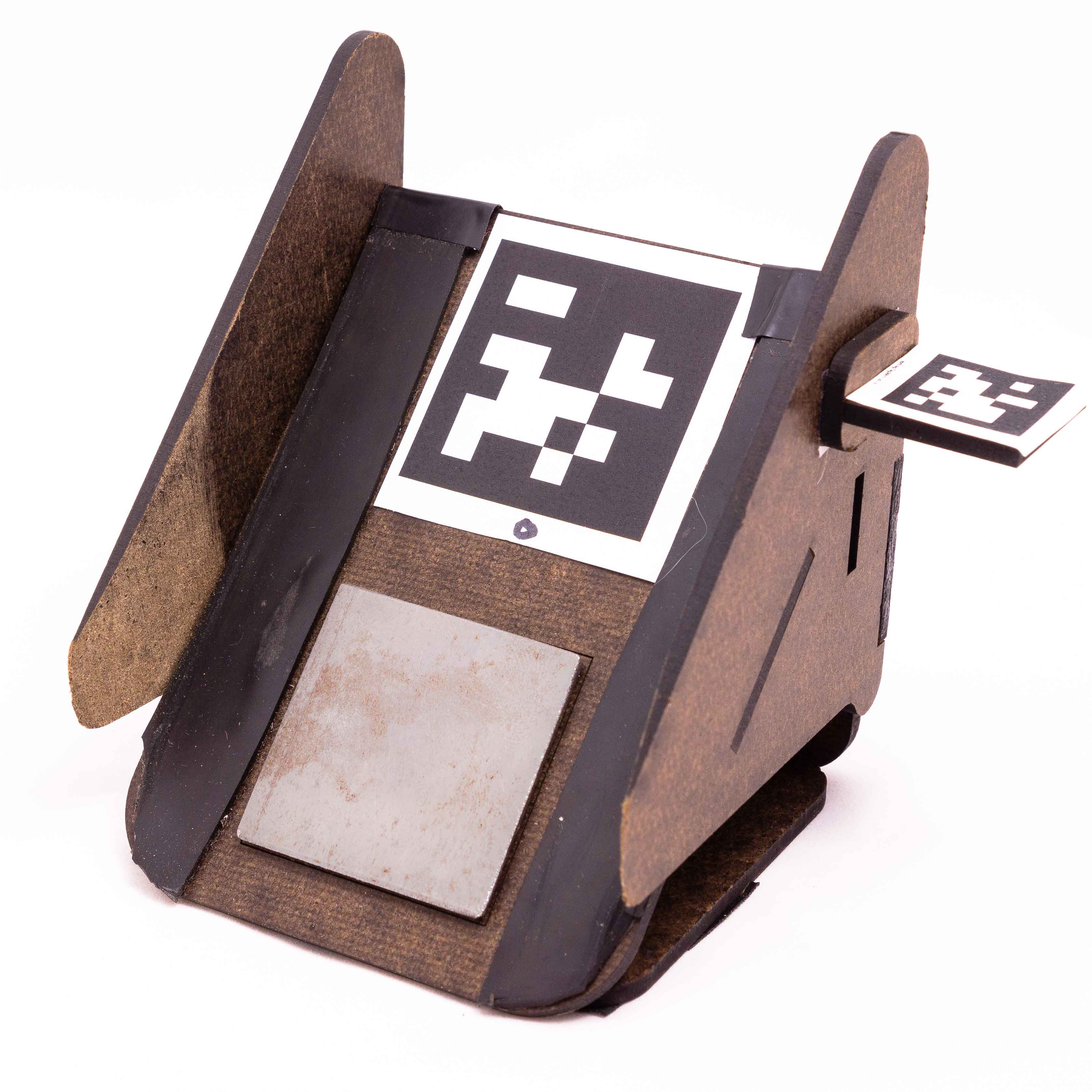}
\includegraphics[width=0.45\columnwidth]{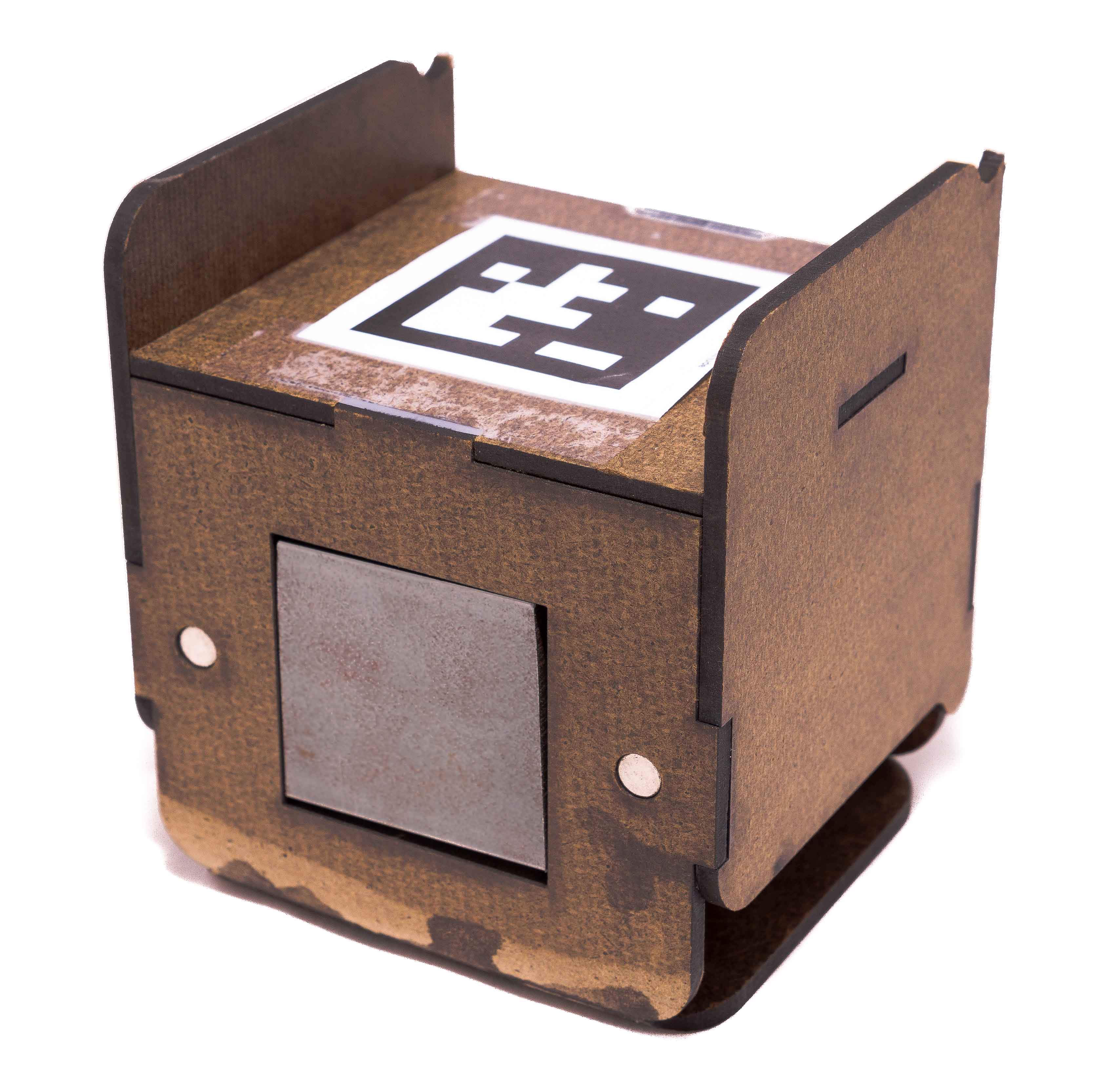}
\caption{Wedge and Block Augmentation Modules}
\label{fig:augmentation_modules} 
\end{center}
\end{figure}

We address this issue by extending the SMORES-EP hardware system with passive elements called \textit{environment augmentation modules}.
We use the \textit{Wedge} and \textit{Block} augmentation modules shown in Figure \ref{fig:augmentation_modules}.
Wedge and block modules are designed to work synergistically with SMORES-EP, providing features that use the best modes of locomotion (driving), manipulation (magnetic attachment), and sensing (AprilTags) available to SMORES-EP.

Blocks are the same size as a module (80mm cube), and wedges are half the size of a block (an equilateral right triangle with two 80mm sides).  Both are made of lightweight laser-cut medium-density fiberboard (blocks are 162g, wedges are 142g) and equipped with a steel attachment point for magnetic grasping.  
Neodymium magnets on the back faces of wedges, and the front and back faces of blocks, form a strong connection in the horizontal direction.
Interlocking features on the top and bottom faces of the blocks, and the bottom faces of the wedges, allow them to be stacked vertically.
Wedges provide a 45-degree incline with a high-friction rubber surface, allowing a set of 3 or more modules to drive up them.  Side walls on both the wedges and blocks ensure that SMORES-EP modules stay aligned to the structure and cannot fall off while driving over it.  The side walls of wedges are tapered to provide a funneling effect as modules drive onto them, making them more tolerant to misalignment. Each wedge and ramp has unique AprilTag fiducials on its faces, allowing easy identification and localization during construction and placement in the environment.

\begin{figure}
\begin{center}
\includegraphics[width=0.367\columnwidth]{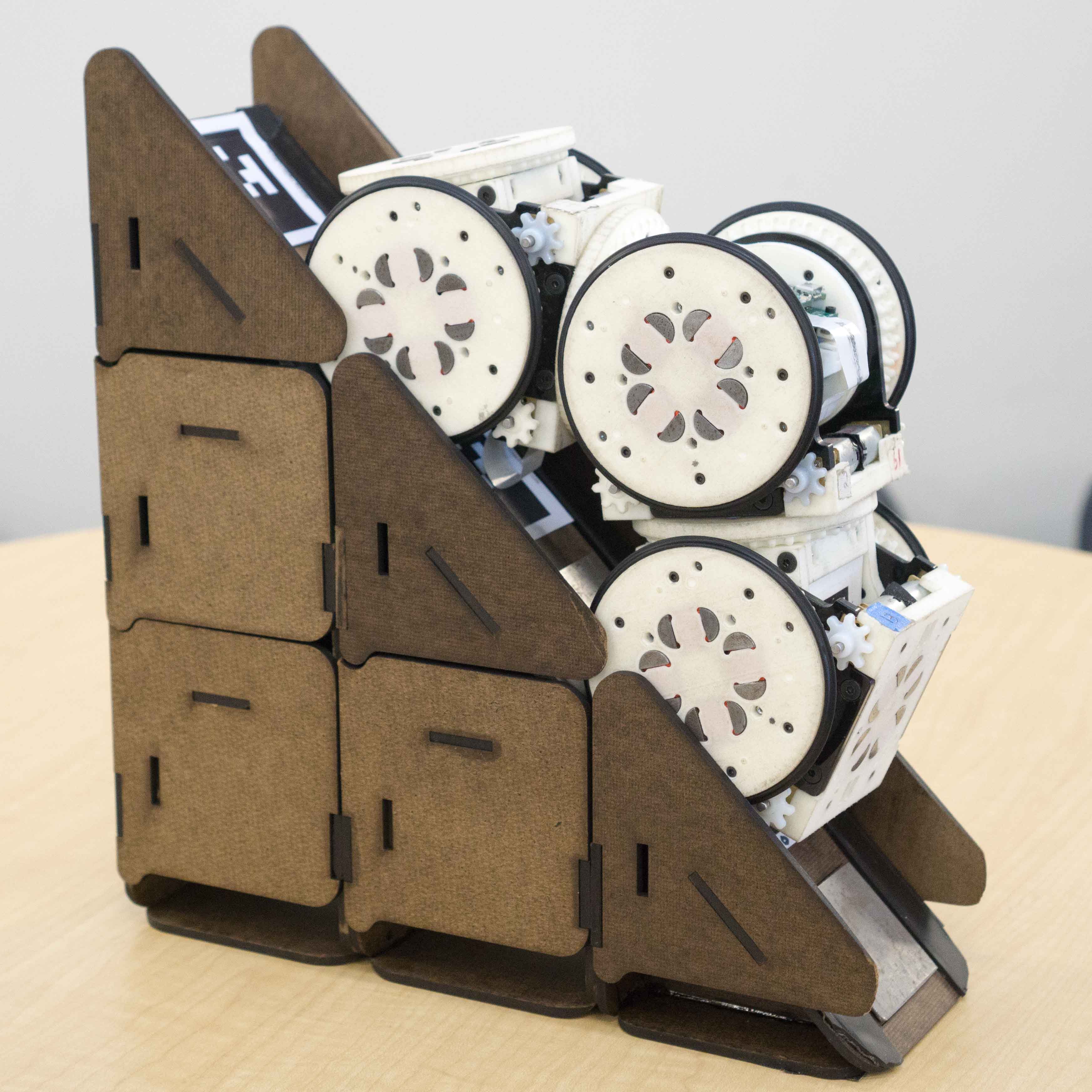}
\includegraphics[width=0.55\columnwidth]{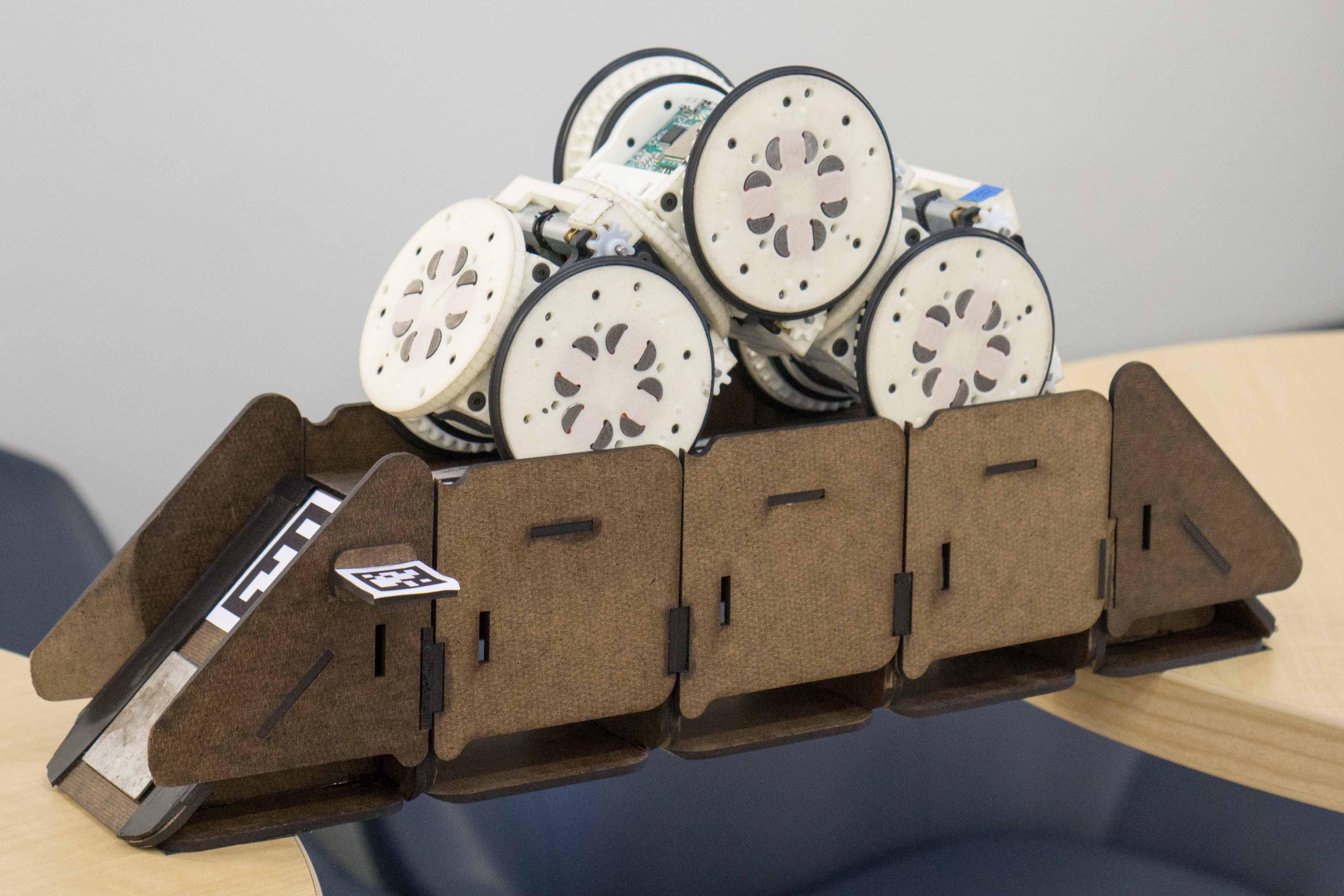}
\caption{Bridge and Ramp}
\label{fig:bridge_and_ramp} 
\end{center}
\end{figure}

Wedges and blocks allow a SMORES-EP cluster to autonomously construct bridges or ramps that allow it to reach higher heights and cross wider gaps than it could with robot modules alone (Figure~\ref{fig:bridge_and_ramp}).  Provided enough time, space, and augmentation modules are available, there is no limit to the height of a ramp that can be built.  Bridges have a maximum length of 480mm (longer bridges cannot support a load of three SMORES-EP modules in the center). 
%
%
\subsection{High-Level Planner}
\label{sec:high-level}

We utilize a high-level planner that allows users to control low-level robot actions by defining tasks at high-level with a formal language \cite{daudelin2017integrated}.
The high-level planner serves two main functions. First it acts as a mission planner, automatically \textit{synthesizing} a robot controller (finite state automaton) from user-given task specifications.
Second, it \textit{executes} the generated controller, commanding the robot to react to the sensed environment and complete the tasks.
In this work, the high-level planner integrates with a \textit{robot design library} of user-created robot configurations and behaviors, as well as a \textit{structure library} of structures that can be deployed to alter the environment.
Users do not explicitly specify configurations and behaviors for each task, but rather define goals and constraints for the robot.
Based on the task specifications, the high-level planner chooses robot configurations and behaviors from the design library, and executes them to satisfy the tasks.  When necessary, the planner will also choose to build a structure from the structure library to facilitate task execution.

\begin{figure}
\begin{center}
\includegraphics[width=\columnwidth]{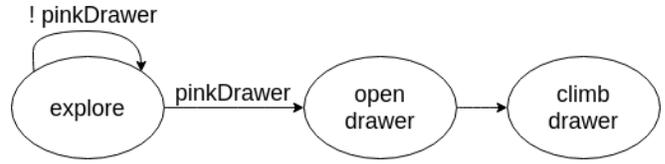}
\caption{An example of synthesized robot controller}
\label{fig:aut}
\end{center}
\vspace{-1em}
\end{figure}

Consider the following example task:  The robot is asked to look for a pink drawer, open the drawer, and then climb on top of it. The mission planer synthesizes the controller shown in Figure~\ref{fig:aut}.  
Each state in the controller is labeled with a desired robot action, and each transition is labeled with perceived environment information; for example, the ``climb drawer'' action is specified to be any behavior from the library with properties \textit{climb} in a \textit{ledge} environment.
In our previous framework \cite{daudelin2017integrated}, the high-level planner could choose to reconfigure the robot whenever needed to satisfy the required properties of the current action and environment.

In this work, the high-level planner can choose not only to change the abilities of the robot (reconfiguration), but also the properties of the environment (environment augmentation).
We expand our framework by introducing a library of structures $S=\lbrace s_1, s_2, \ldots, s_N \rbrace$, where each structure is defined as $s_n = \lbrace \mathcal{F}_n, A_n \rbrace$. $\mathcal{F}_n$ is an environment feature template that specifies the kind of environment the structure can augment, and which can be identified by the environment characterization algorithm described in Section~\ref{sec:characterization}.  The \textit{assembly plan} $A_n$ is itself a high-level task controller (finite state automaton), specifying the required building blocks needed to create the structure and the order in which they may be assembled.  As with other tasks in our framework, construction actions within assembly plans are specified in terms of behavior properties (e.g. \textit{pickUpBlock}, \textit{placeWedge}) that the high-level planner maps to appropriate configurations and behaviors from the robot design library. 

For the example in Figure~\ref{fig:aut}, if no behavior in the library satisfies the ``climb drawer'' action, the high-level planner will consider augmenting its environment with a structure.  It passes a set of feature templates to the environment characterization subsystem, which returns a list of matched features (if any are found), as well as two lists of regions $R^1=\{r_0^1, r_1^1,\dots\},\ R^2=\{r_0^2, r_1^2,\dots\}$ that the matched features connect.

To decide what structure to build, the high-level planner considers the available augmentation modules in the current environment, the current robot configuration, and the distance from the structure build-point to the robot goal position.  After selecting a structure, the high-level planner executes its assembly plan to construct it.  Once the structure is built, the high-level planner considers regions $r^1_i$ and $r^2_i$ to be  connected and traversable by the robot, allowing it to complete its overall task of climbing onto the drawer. 

%

\begin{figure}
\begin{center}
\includegraphics[width=\columnwidth]{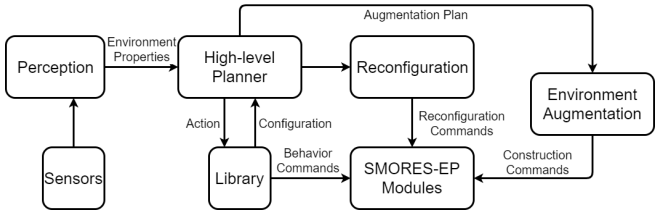}
\caption{System Overview Flowchart}
\label{fig:overview}
\end{center}
\vspace{-1cm}
\end{figure}
                       
\section{System Integration}
\label{sec:integration}
We integrate our environment augmentation tools into the system introduced in \cite{daudelin2017integrated}, as shown in Figure~\ref{fig:overview}.
The high-level planner automatically converts user defined task specifications to controllers from a robot design library.
It executes the controller by reacting to the sensed environment, running appropriate behaviors from the design library to control a set of hardware robot modules. 
Active perception components perform simultaneous localization and mapping (SLAM), and characterize the environment in terms of robot capabilities.
Whenever required, the reconfiguration subsystem controls the robot to change configurations.

The system used the Robot Operating System (ROS)\footnote{http://www.ros.org} for a software framework, networking, and navigation. SLAM was performed using RTAB-MAP\cite{rtabmap}, and color detection was done using CMVision\footnote{CMVision: http://www.cs.cmu.edu/$\sim$jbruce/cmvision/}.

\begin{figure}
\begin{center}
\includegraphics[width=0.3\textwidth]{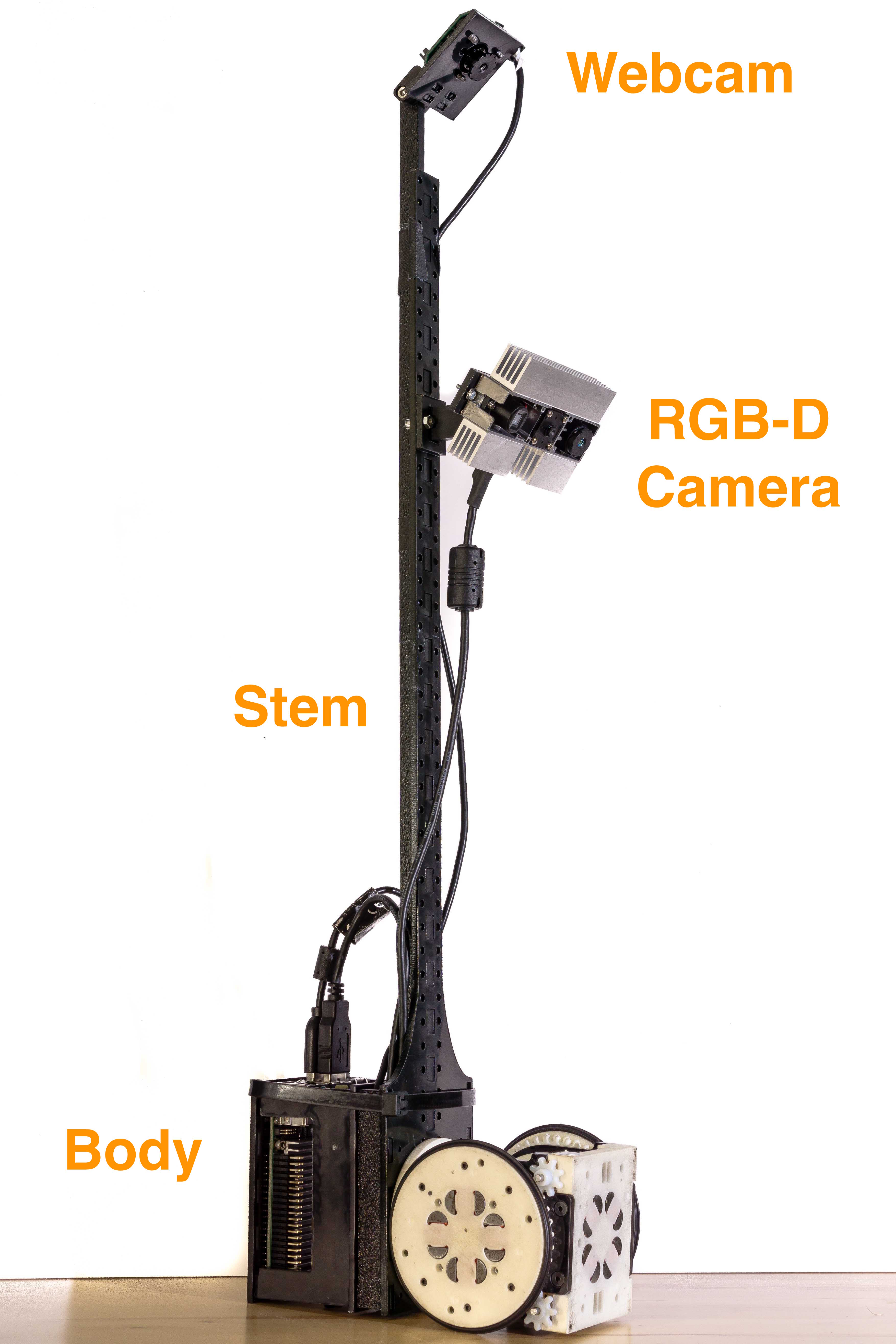}
\caption{Sensor Module with labelled components.  UP board and battery are inside the body.}
\label{fig:sensor-module}
\end{center}
\vspace{-1.4cm}
\end{figure}

SMORES-EP modules have no sensors that allow them to gather information about their environment. To enable autonomous operation, we use the \textit{sensor module} shown in Figure~\ref{fig:sensor-module}.
The sensor module has a 90mm $\times$ 70mm $\times$ 70mm body with thin steel plates on its front and back that allow SMORES-EP modules to connect to it. Computation is provided by an UP computing board with an Intel Atom 1.92 GHz processor, 4 GB memory, and a 64 GB hard drive. A USB WiFi adapter provides network connectivity. A front-facing Orbecc Astra Mini RGB-D camera enables the robot to map and explore its environment and recognize objects of interest.  A thin stem extends 40cm above the body, supporting a downward-facing webcam. This camera provides a view of a  1m $\times$ 0.75m area in front of the sensor module, and is used to track AprilTag \cite{olson2011apriltag} fiducials on modules and augmentation modules for reconfiguration and structure building. A 7.4V, 2200mAh LiPo battery provides about one hour of running time.
%
%
%
\section{Experiment Results}
\label{sec:experiments}
\begin{figure*}
\begin{center}
\begin{tabular}{c c c}
\includegraphics[width=0.3\textwidth]{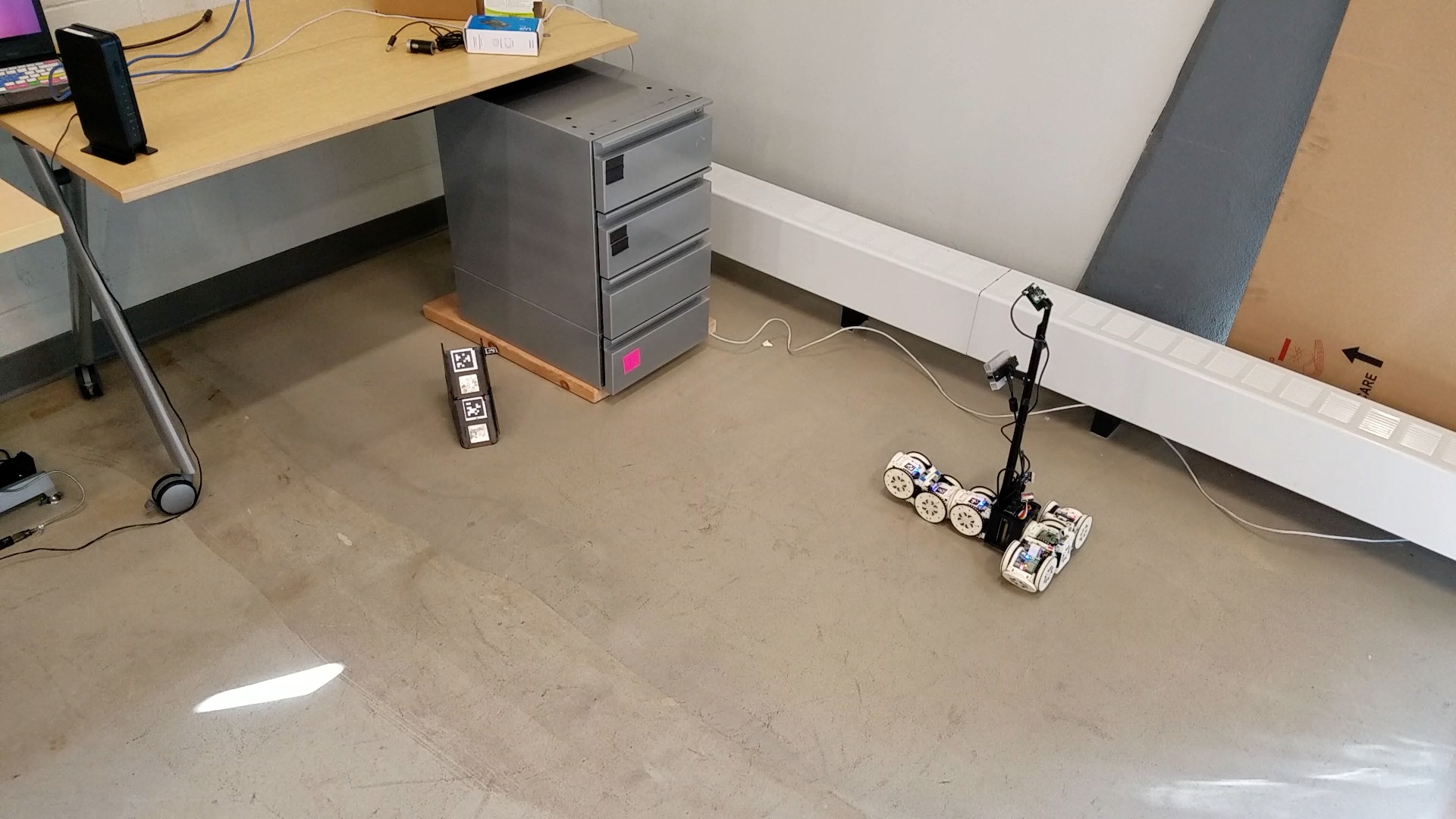} &
\includegraphics[width=0.3\textwidth]{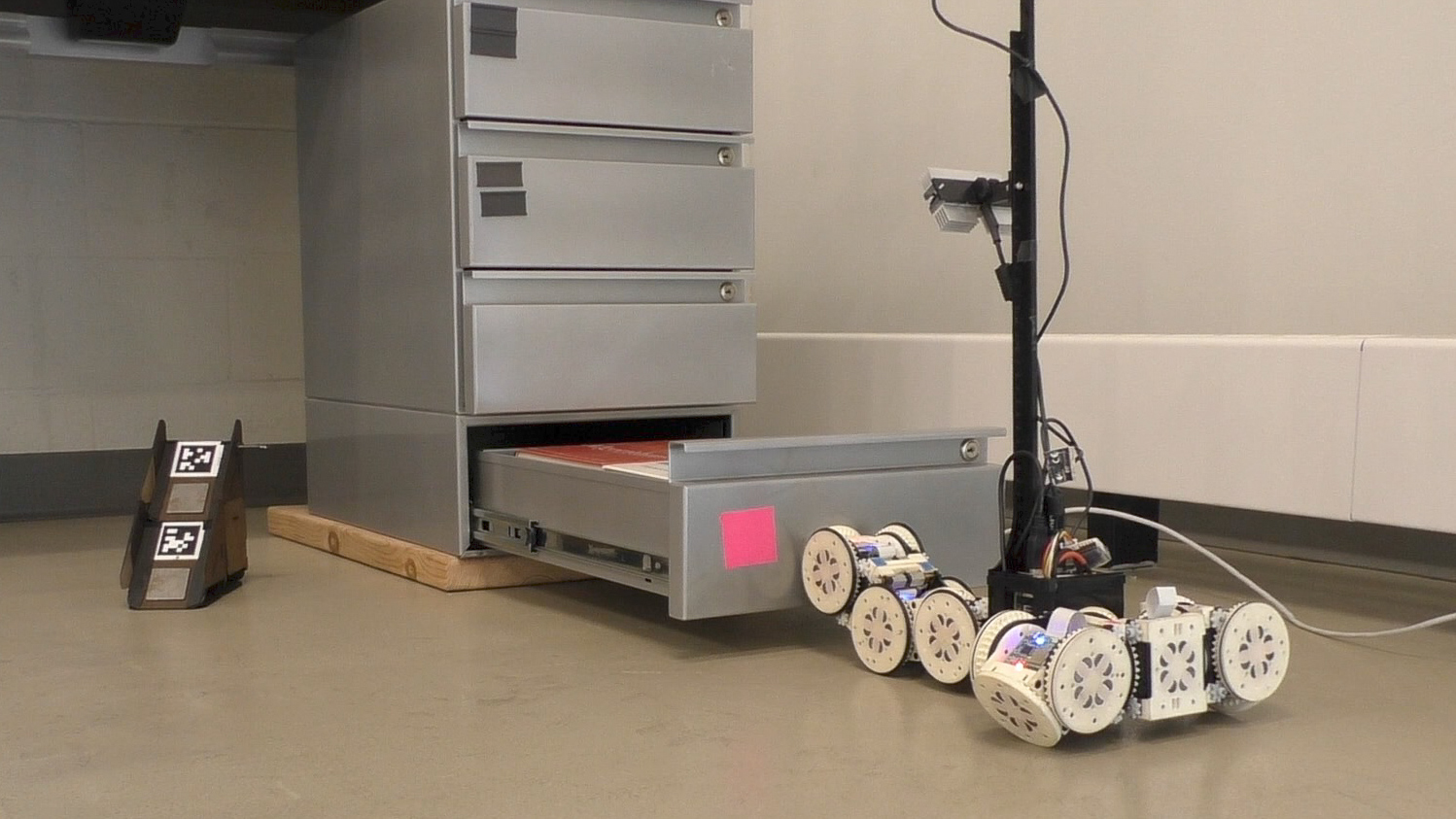} &
\includegraphics[width=0.3\textwidth]{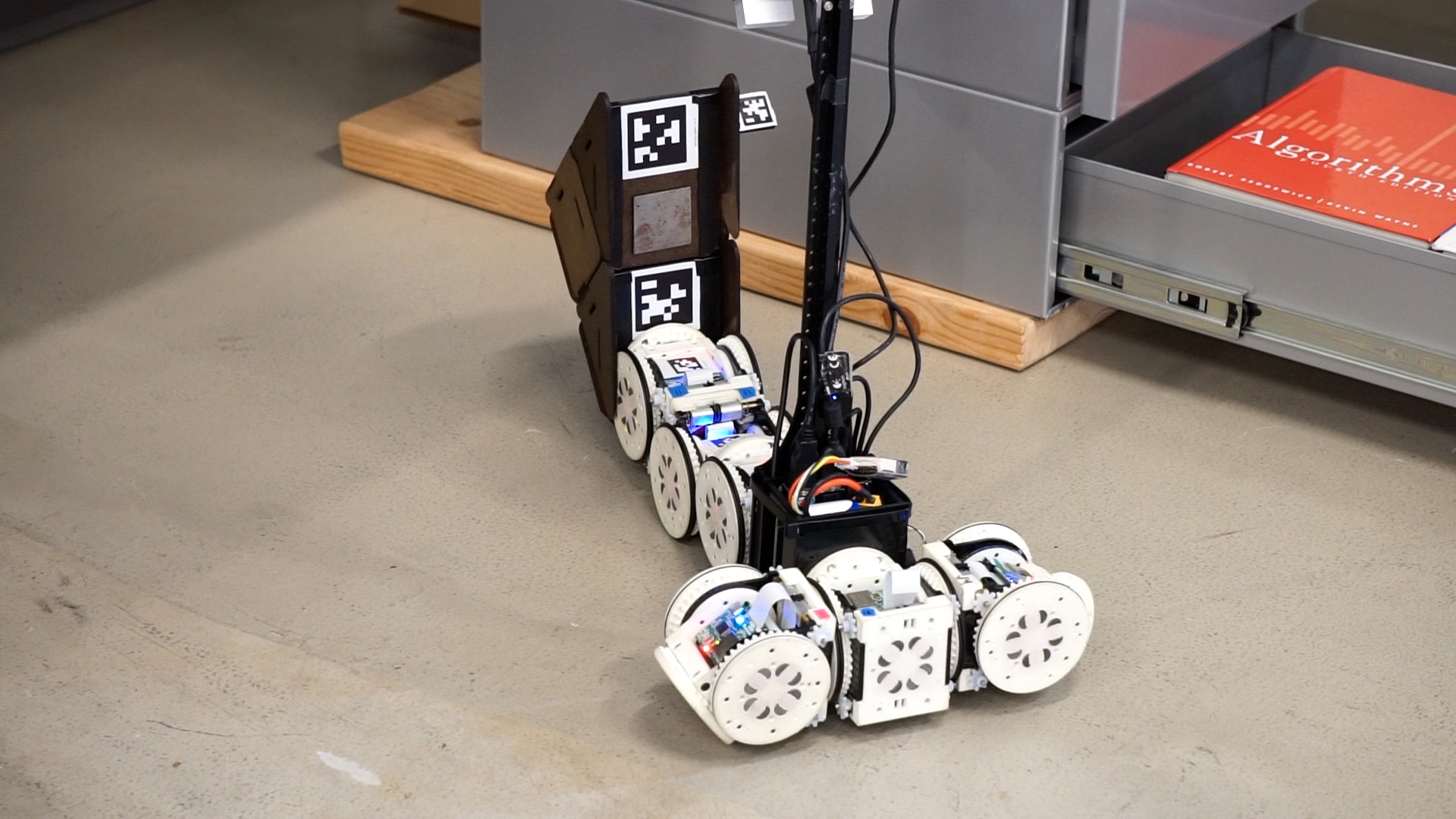} 
\\
\includegraphics[width=0.3\textwidth]{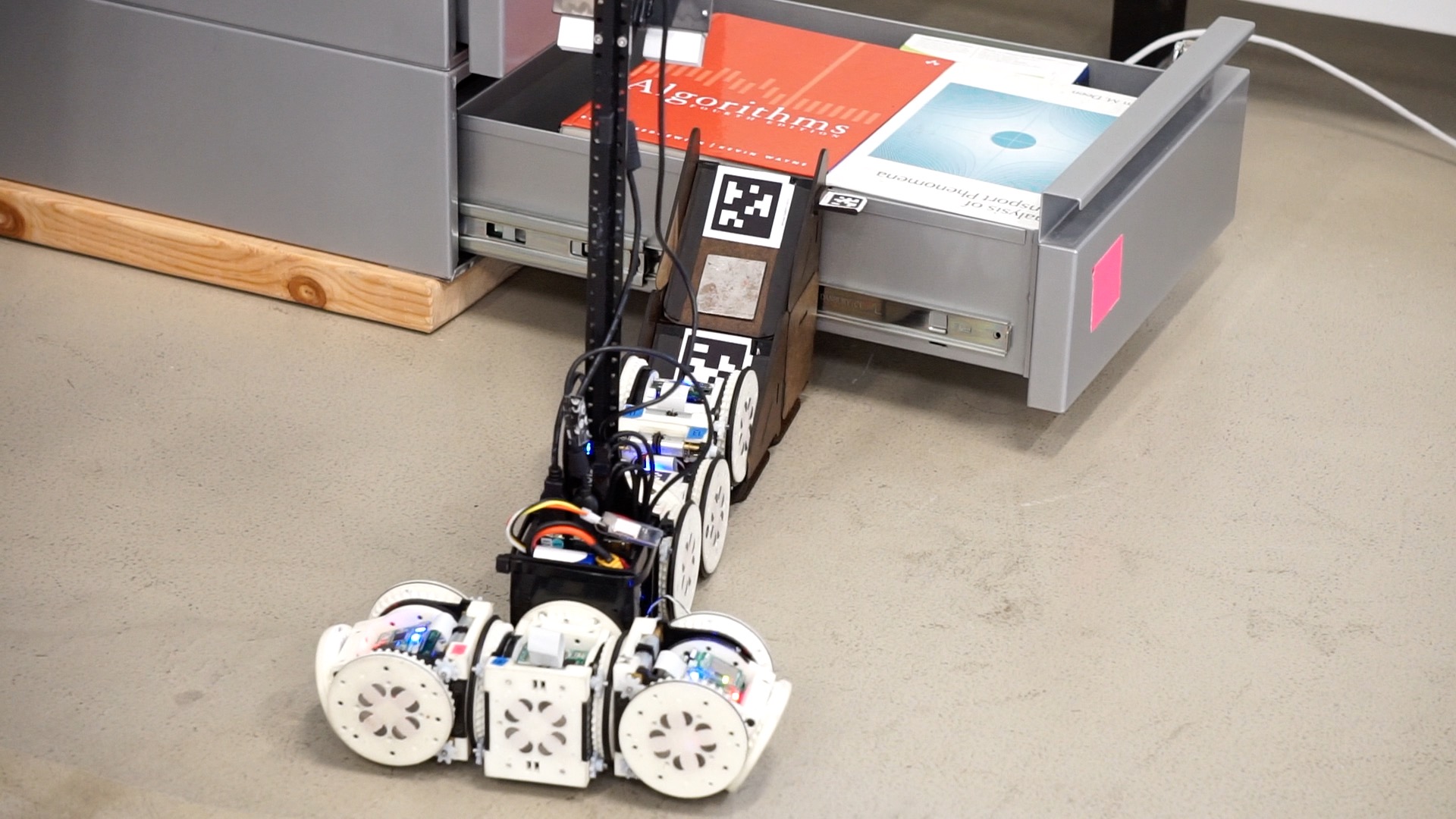} & 
\includegraphics[width=0.3\textwidth]{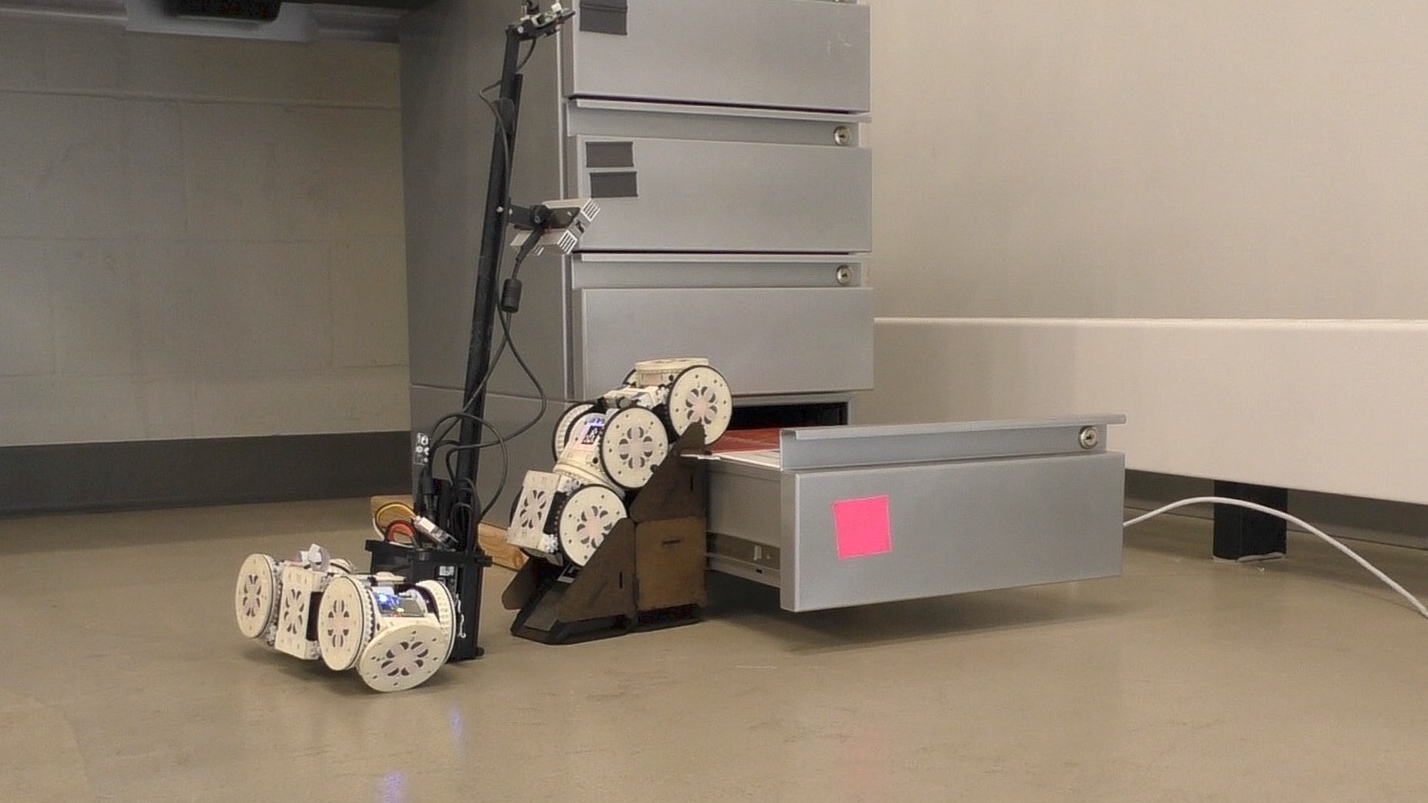} & 
\includegraphics[width=0.3\textwidth]{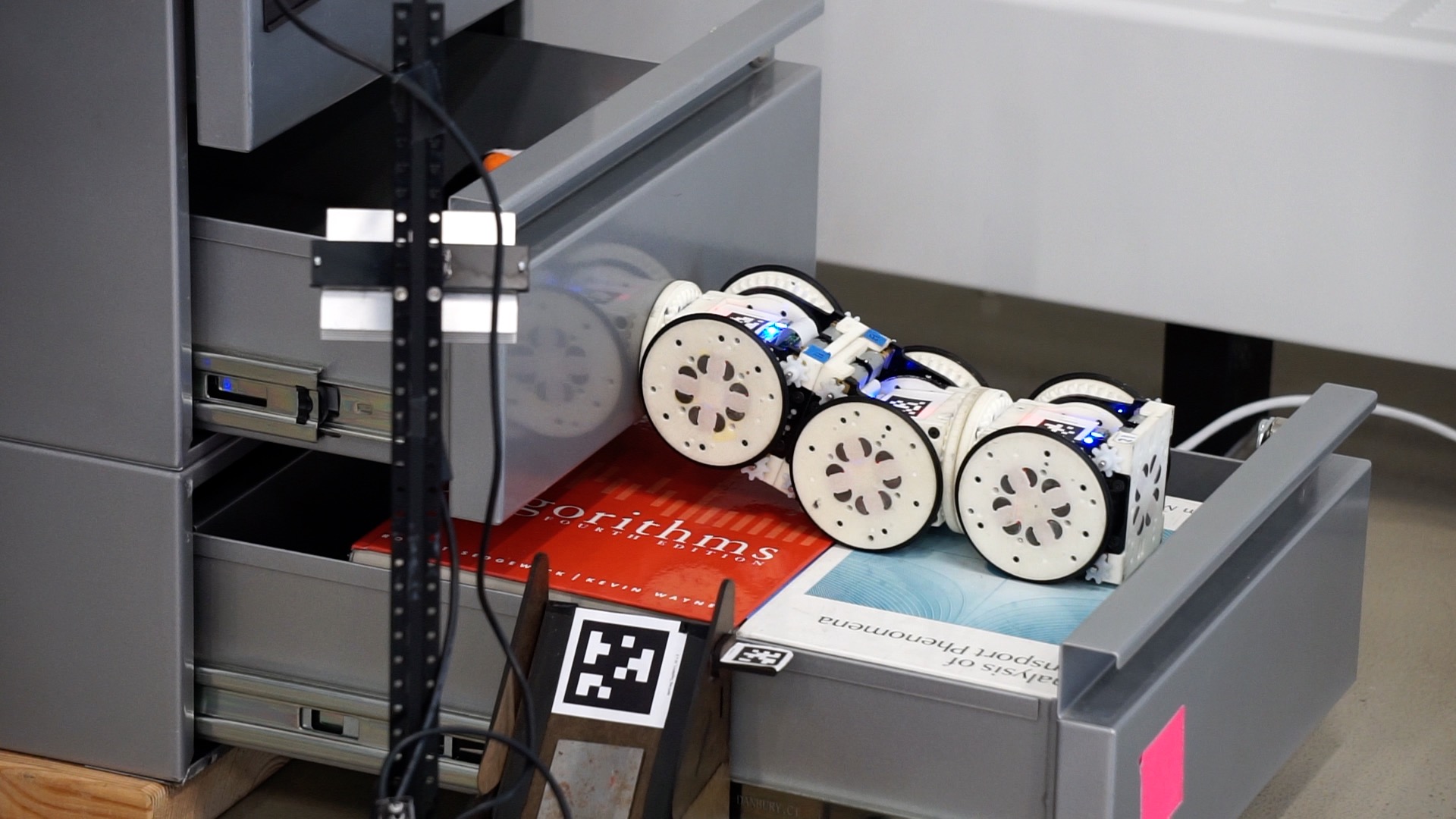}
\end{tabular}
\caption{Snapshots throughout Experiment I. From left to right, top to bottom: i) Experiment start ii) Opening first drawer iii) Picking up ramp iv) Placing ramp next to open drawer. v) Reconfiguring and climbing ramp vi) Opening second drawer}
\vspace{-0.7cm}
\label{fig:experiment_1} 
\end{center}
\end{figure*}

Our system can generalize to arbitrary environment augmentations and high-level tasks. We validate our system in two hardware experiments that require the same system to perform tasks requiring very different environment augmentations for successful completion. In both experiments, the robot autonomously perceives and characterizes each environment, and synthesizes reactive controllers to accomplish the task based on the environment. Videos of the full experiments accompany this paper, and are also available online at \url{https://youtu.be/oo4CWKWyfvQ}.

\subsection{Experiment I}
For the first experiment, the robot is tasked with inspecting two metal desk drawers in an office.
It is asked to locate the set of drawers (identified with a pink label), open the two lowest drawers, and inspect their contents with its camera.
The robot can open metal drawers by magnetically attaching to them and pulling backwards, but it is unable to reach the second drawer from the ground. Therefore, it can only open the second drawer if it can first open the bottom drawer and then climb on top of the things inside it.

Figure \ref{fig:experiment_1} shows snapshots throughout the robot's autonomous performance of Experiment I. After recognizing and opening the first drawer, the robot characterizes the environment with the opened drawer and identifies the side of the drawer as a ``ledge'' feature. The high-level planner recognizes that the ledge is too high for the current configuration to climb, and furthermore that there is no other configuration in the library to which the robot can transform that could climb the ledge, leaving
environment augmentation as  the only strategy that can complete the task.
Observing a ramp structure in the environment, the high-level planner commands the robot to acquire the ramp, place it at the ``ledge'' feature detected by the characterization algorithm, climb the drawer, and complete the mission.


In a second version of the same experiment, the first drawer is empty. When the robot characterizes the environment containing the drawer, it identifies no ``ledge'' features, since the drawer no longer matches the requirements of the feature.
As a result, it recognizes that environment augmentation is not possible, and the mission cannot be completed.
%
\subsection{Experiment II}
%
The environment for Experiment II consists of two tables separated by a 16 cm gap. The robot begins the experiment on the left table with two wedges and one block. To complete its mission, the robot must cross the gap to reach a pink destination zone on the right table.

\begin{figure*}
\begin{center}
\begin{tabular}{c c c}
\includegraphics[width=0.3\textwidth]{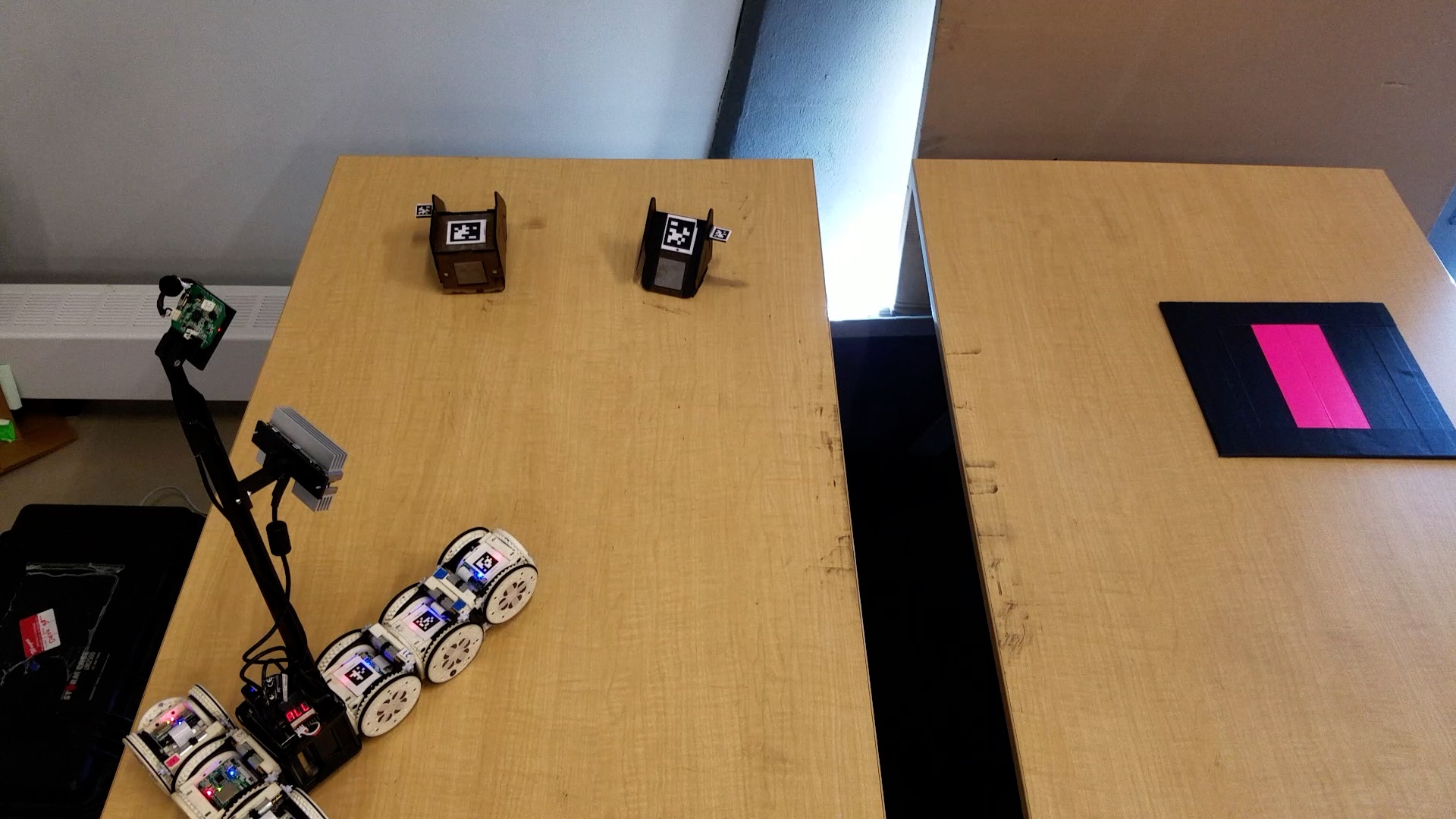} &
\includegraphics[width=0.3\textwidth]{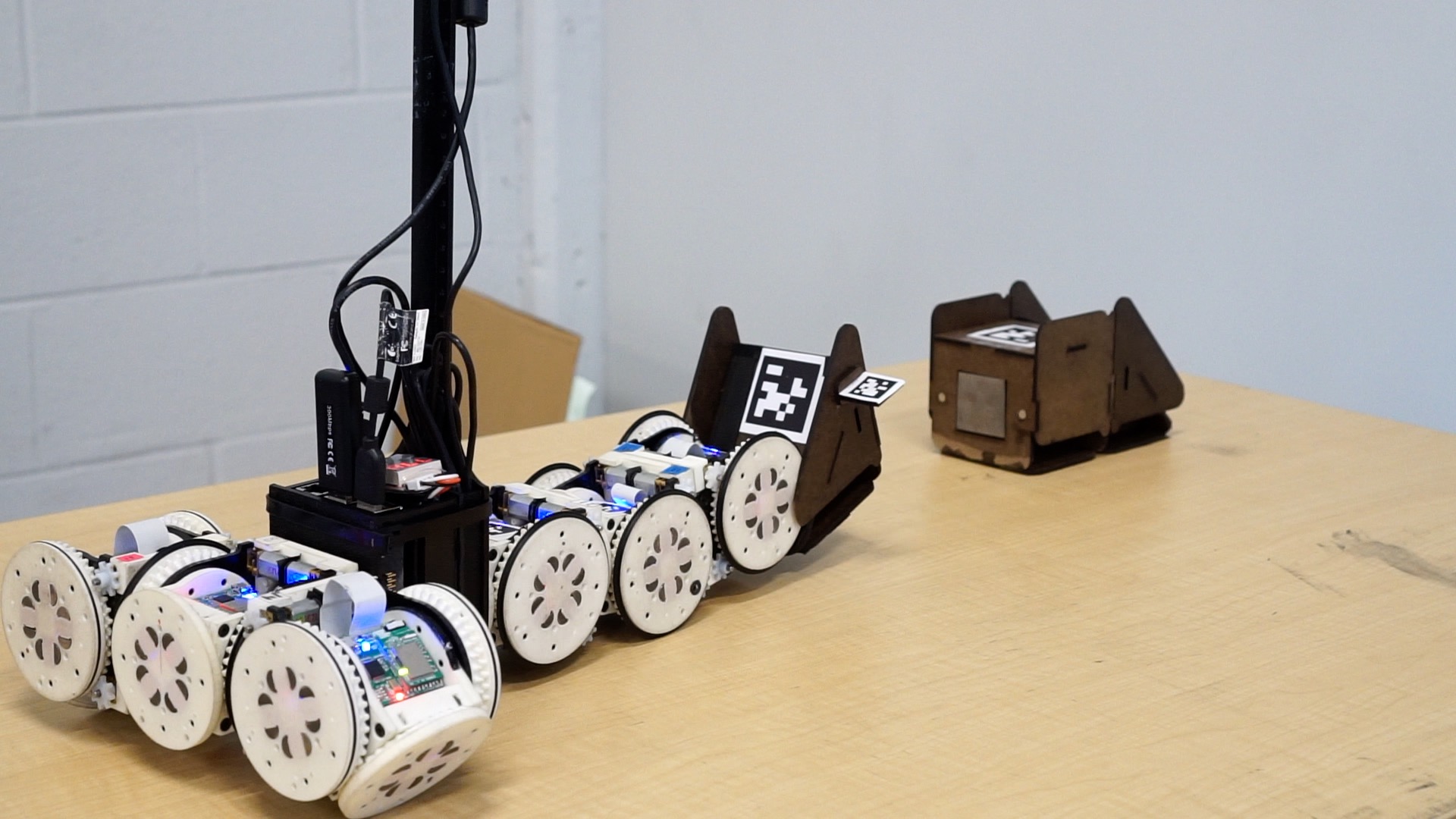} &
\includegraphics[width=0.3\textwidth]{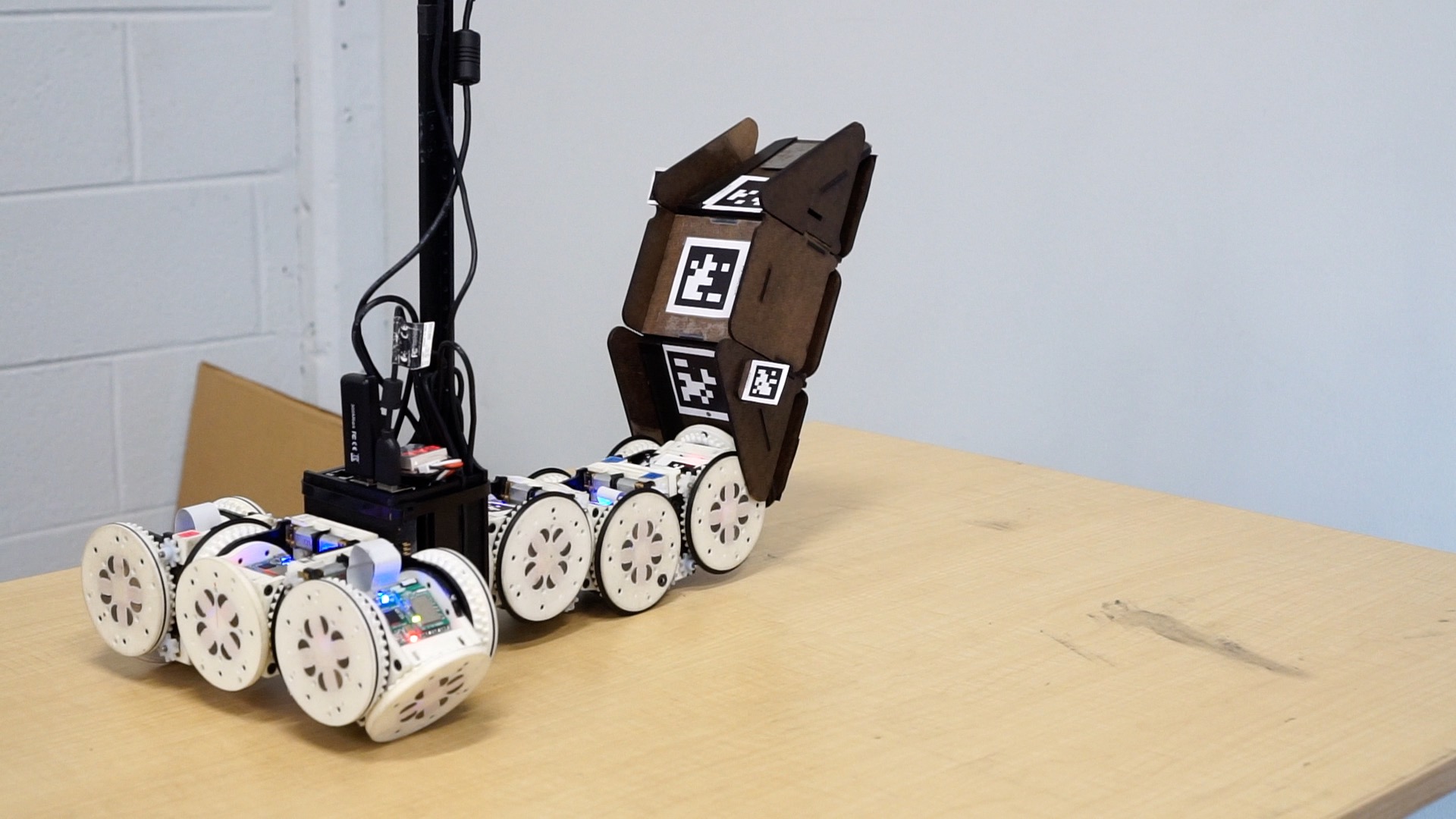}
\\
\includegraphics[width=0.3\textwidth]{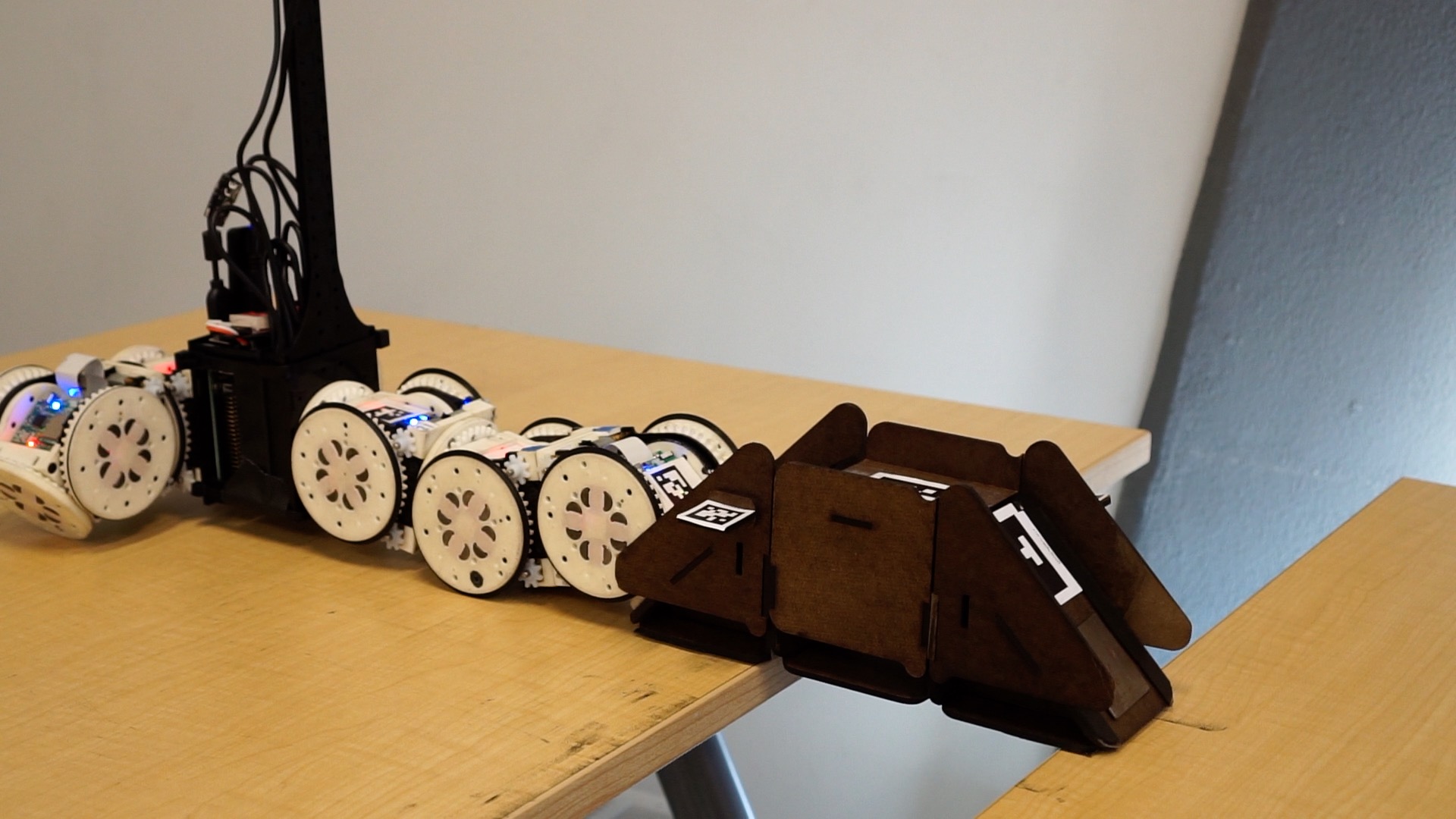} & 
\includegraphics[width=0.3\textwidth]{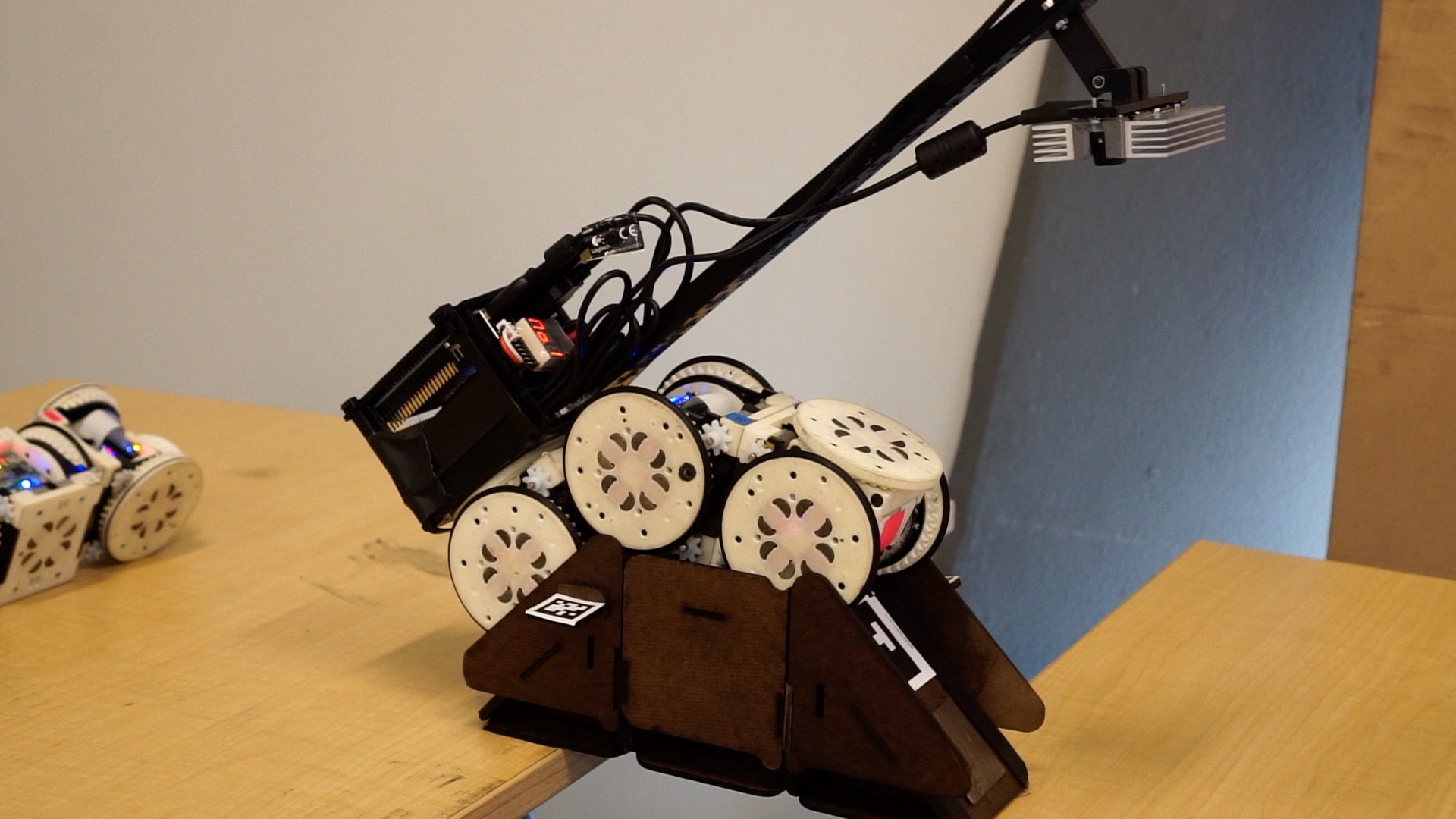} &
\includegraphics[width=0.3\textwidth]{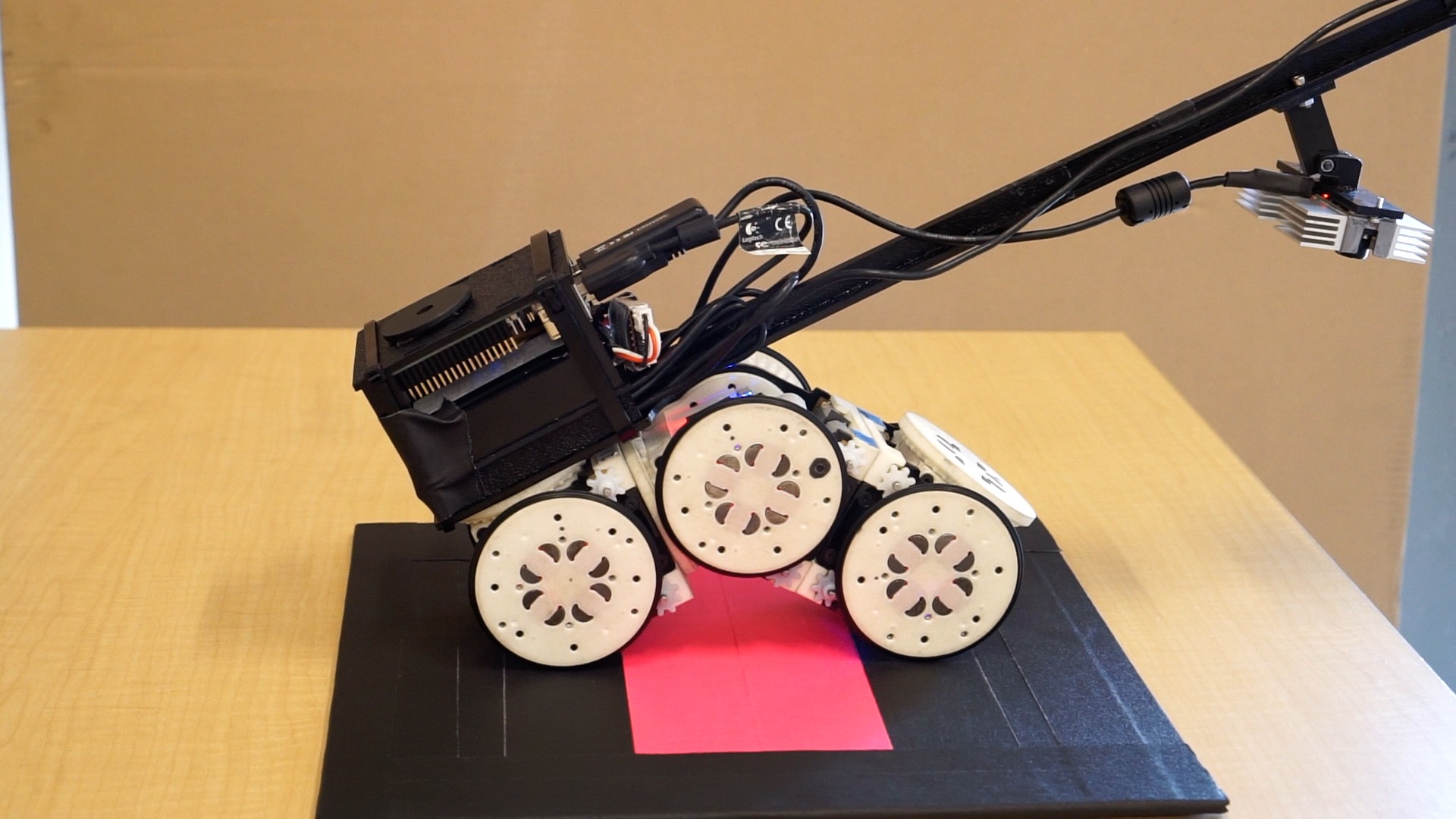}
\\

\end{tabular}
\caption{Snapshots throughout Experiment II. From left to right, top to bottom: i) Experiment start ii) Assembling bridge iii) Transporting bridge iv) Placing bridge over gap. v) Reconfigure and cross bridge. vi) Arrive at the target zone.}
\vspace{-0.6cm}
\label{fig:experiment_2} 
\end{center}
\end{figure*}

Figure \ref{fig:experiment_2} shows snapshots throughout Experiment II. This time, characterization of the environment identifies that the pink goal zone is in a separate region from the robot, and also identifies several ``gap'' features separating the two regions.
Recognizing that the gap is too wide for any configuration in the design library to cross unassisted, the high-level planner concludes it must build a bridge across the gap to complete its mission. It begins searching for materials, and quickly identifies the three available augmentation modules, which it autonomously assembles into a bridge.  It then places the bridge across the gap and crosses to complete its mission.
%
%
\section{Discussion}
\label{sec:discussion}
%
Block and wedge modules demonstrably expand the physical capabilities of SMORES-EP, allowing the system to climb to a high ledge and cross a wide gap to complete tasks that would have been very difficult with the SMORES-EP modules alone. Perception tools accurately characterize augmentable features in the environment.  High-level reasoning tools identify when environment augmentation is necessary to complete a high-level task, and reactively sequence locomotion, manipulation, construction, and reconfiguration actions to accomplish the mission.
The presented work represents the first time that modular robots have successfully augmented their environment by deploying passive structures to perform high-level tasks.

As with our previous work with autonomous modular robots \cite{daudelin2017integrated}, robustness proved challenging. Out of 8 test runs of Experiment I, the robot successfully completed the entire task once. Table~\ref{table:errors} shows the outcomes of 8 runs of each experiment. The largest source of error was due to hardware failures such as slight encoder mis-calibration or wireless communication failure.
Creating more robust hardware for modular robots is challenging due to the constrained size of each module, and the higher probability of failure from higher numbers of components in the system.

Perception-related errors were another frequent cause of failure. These were due in part to mis-detections by the characterization algorithm, or because the accuracy in finding location and orientation of features was not high enough for the margin of error of the robot when placing structures. Finally, navigation failures occurred throughout development and experiments due to cumulative SLAM errors as the robot navigates the environment. We found that it was important to minimize in-place rotation of the robot, and to avoid areas without many features for visual odometry to use.

%
\begin{table}
\centering
\begin{tabular}{|c|c c|c c|}
\hline
\textbf{Outcome} & \multicolumn{2}{c|}{\textbf{Exp 1}} &  \multicolumn{2}{c|}{\textbf{Exp 2}}  \\ 
\hline
Success & 2 & (25.0\%) & 3 & (37.5\%) \\
\hline
Perception-Related Failure & 2 & (25.0\%)  & 2 & (25.0\%) \\
\hline
Navigation Failure & 1 & (12.5\%) & 0 & (0.0\%) \\ 
\hline
Hardware Failure & 3 & (37.5\%) & 2 & (25.0\%) \\ 
\hline
Setup Error & 0 & (0.0\%) & 1 & (12.5\%) \\
\hline
\end{tabular}  
\caption{Outcomes for Experiments 1 and 2}
\vspace{-1cm}
\label{table:errors}
\end{table}

%
%
\subsection{Future}
%

In the interest of establishing the deployment of structures as an effective
means to address high-level tasks, this work does not focus on the speed
or scale of construction, demonstrating the use of only small structures
(with three elements).
Future work might attempt to accelerate construction, build larger structures, and attempt larger-scale tasks with SMORES-EP.  For the purposes of this work, structure assembly plans ($A_n$) were create manually, but this process could be automated by employing established assembly planning algorithms \cite{Seo2013,Werfel2007}.
Assembly might be significantly accelerated by using multiple builders in parallel, as some other collective robot construction systems have done \cite{petersen2011termes}.  To truly scale to large tasks, a large number of block and wedge modules must be available in the environment, or better, autonomously transported into the environment.  Developing mechanisms for transporting building material to a task location remains an open challenge for future work.

While our system implementation is tightly coupled to the SMORES-EP hardware, the concepts, system architecture, and theoretical frameworks could be applied widely.  In particular, most elements of the framework could be directly applied to the Termes \cite{petersen2011termes} or foam-ramp building robots \cite{napp2014distributed} that have similar construction and locomotion capabilities to SMORES-EP, provided that appropriate sensing and perception capabilities were established.
\subsection{Conclusion}
To conclude, this paper presents tools that allow a modular robot to autonomously deploy passive structures as a means to complete high-level tasks involving locomotion, manipulation, and reconfiguration. This work expands the physical capabilities of the SMORES-EP modular robot, and extends our existing frameworks for addressing high-level tasks with modular robots by allowing both the robot morphology and the environment to be altered if doing so allows the task to be completed.  We validate our system in two hardware experiments that demonstrate how the hardware, perception tools, and high-level planner work together to complete high-level tasks through environment augmentation.
\section*{Acknowledgments}
This work was funded by NSF grant numbers CNS-1329620 and CNS-1329692.


\bibliographystyle{IEEEtran}
\bibliography{references}
\end{document}